\documentclass[12pt]{iopart}

\usepackage{iopams}  

\usepackage{cite}
\usepackage{graphicx}
\usepackage{bm}

\usepackage{color}

\begin{document}

\title{Embedding stochastic differential equations into neural networks via dual processes}

\author{Naoki Sugishita and Jun Ohkubo}
\address{Graduate School of Science and Engineering, Saitama University, \\
255 Shimo-Okubo, Sakura-ku, Saitama 338-8570, Japan}
\ead{johkubo@mail.saitama-u.ac.jp}
\vspace{10pt}
\begin{indented}
\item[]
\end{indented}

\begin{abstract}
We propose a new approach to constructing a neural network for predicting expectations of stochastic differential equations. The proposed method does not need data sets of inputs and outputs; instead, the information obtained from the time-evolution equations, i.e., the corresponding dual process, is directly compared with the weights in the neural network. As a demonstration, we construct neural networks for the Ornstein-Uhlenbeck process and the noisy van der Pol system. The remarkable feature of learned networks with the proposed method is the accuracy of inputs near the origin. Hence, it would be possible to avoid the overfitting problem because the learned network does not depend on training data sets.
\end{abstract}

%\maketitle

\section{\label{sec:intro}Introduction}

Many researchers have studied the evolution of dynamical or stochastic systems in physics and other research fields. Recently, the amount of data has exponentially increased, and there are many studies on dynamical systems based on these large data sets. One of the aims of these studies is to transform observed data into predictive models of the physical world, and neural networks are a hopeful candidate for this aim. Of course, a simple application of conventional neural networks would not be enough because the physical world has many characteristics due to various constraints. For example, it would be beneficial to equip features of the time-evolution into the learning steps to make the prediction more accurate. Some ideas have appeared recently; the numerical scheme for time-evolution with multi-step time-stepping schemes is employed \cite{Raissi2018}. Some works focused on network architectures. Reference \cite{Long2018} discussed a network architecture for the time-evolution partial differential equation. In \cite{Wong2018, Wu2019}, recurrent neural networks were applied for model predictive control. There is a study to deal with fluid flow simulation with long short-term memory (LSTM) \cite{Wiewel2019}. Other types of discussions based on the universal approximation theorem were given in \cite{Lu2021}, in which DeepONet was also proposed.

The physics-informed machine learning has the same research direction. In \cite{Wu2018, Raissi2019}, frameworks for learning partial differential equations were discussed; see a review \cite{Karniadakis2021} for this topic. There are studies in which conservation laws are combined with learning; the inclusion of the Hamiltonian structures makes learning more stable \cite{Greydanus2019}. A recent paper gives a good review of this topic in the introduction \cite{Mattheakis2022}. However, compared with the studies on deterministic dynamical systems, there has been little discussion about systems with noise, i.e., stochastic systems. Such stochastic systems are the topic of the present paper.

One of the problems of machine learning for physical subjects is the data sets. For example, when one constructs a neural network for prediction from an input coordinate, it is necessary to prepare data with various initial conditions. Note that stochastic systems require large data sets compared with deterministic systems; in stochastic systems, we must consider statistics of predictions, and the calculation of expected values takes high computational cost. Hence, we need more computational effort in the data preparation and the learning steps. However, if we have the information of equations governing the system as the prior knowledge, it could be possible to reduce the size of the data sets and costs for learning. Of course, it will be possible to apply additional online learning steps to the trained network to obtain a more accurate one. Hence, the key question of the present paper is as follows: How should we cooperate with the information on the time-evolution equation for stochastic systems? If we can embed the prior knowledge directly without data sets, the method would complement the conventional ones.

\begin{figure}
  \centering
  \includegraphics[width=120mm]{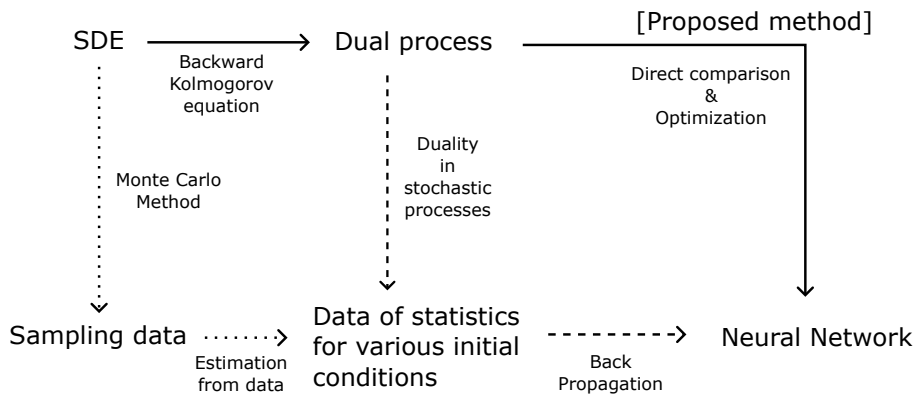}
  \caption{\label{fig:concept} 
Conventional and proposed methods for the learning of stochastic systems. A most naive approach is to generate data with Monte Carlo methods; target statistics are evaluated from the sampled data sets. Employing dual processes enables us to skip the sampling step; it is possible to evaluate the statistics directly by solving the backward Kolmogorov equation. These two approaches give a data set with the inputs and outputs, and then neural networks are learned with conventional backpropagation. By contrast, the proposed method employs direct comparison with the information of the dual processes.
}
\end{figure}

In the present paper, the aim is to predict statistics after time evolution in stochastic systems, especially stochastic differential equations. For this aim, we propose a new method to learn neural networks without generating data sets for statistics. The key of the proposal is the usage of dual processes. The duality of stochastic processes has long been studied in statistical physics and mathematical physics \cite{Liggett2005}, and recent studies clarified that the conventional backward Kolmogorov equation gives dual processes straightforwardly. Figure~\ref{fig:concept} summarizes the proposal. In a most naive approach, we generate data with Monte Carlo samplings and calculate target statistics for various initial conditions. As shown later, one can evaluate the target statistics directly from the dual process. 
In both methods, after obtaining a data set with pairs of initial conditions and the target statistics, the backpropagation gives leaned neural networks. By contrast, the proposed method employs direct comparison with the information of the dual processes. Hence, we utilize an optimization procedure instead of backpropagation. Since there is no need to evaluate statistics from the samplings, it would be possible to write that the proposed method embeds stochastic differential equations directly into neural networks. We demonstrate the proposed method with one-dimensional and two-dimensional noisy systems, which will clarify the learned features different from the conventional approach.

The remaining part of the present paper proceeds as follows. In Section~2, we review the method to evaluate statistics of the stochastic differential equations without any sampling. Section~3 gives the main proposal to construct neural networks directly from the stochastic differential equations. Two numerical demonstrations are given in section~4. Finally, section~5 concludes this paper.

\section{Numerical method to evaluate statistics without sampling}

In the present paper, we only focus on the stochastic differential equations and the statistics after the time-evolution. As for the basics of the stochastic differential equations, see \cite{Gardiner2009} for example. Here, Let $\bm{x}(t)$ be a $D$-dimensional random vector which obeys the following stochastic differential equation:
\begin{eqnarray} \label{eq:sde}
  d\bm{x} = \bm{a}(\bm{x})dt + B(\bm{x})d\bm{W}(t),
\end{eqnarray}
where $\bm{a}(\bm{x})$ is a vector function called the drift coefficient and $B(\bm{x})$ is a matrix-valued function called the diffusion coefficient. $\bm{W}(t)$ is a vector of Wiener processes.

Since one cannot generally solve stochastic differential equations analytically, the Monte Carlo method is employed to sample trajectories from stochastic differential equations. The famous method is the Euler-Maruyama method; for example, see \cite{Kloeden_book}. In the Euler-Maruyama method, a time-discretization is necessary; the sampling steps are time-consuming when we want to evaluate statistics with high accuracy for various initial conditions.

As stated in the introduction, the usage of duality in stochastic processes has long been studied in physics \cite{Liggett2005}. While the usage was mainly restricted to exactly solvable cases, recent studies give a simple derivation of dual processes from stochastic differential equations \cite{Ohkubo2019}; there is also an algorithm based on combinatorics to evaluate the statistics of stochastic differential equations \cite{Ohkubo2022}. Here, there is no need to know the details of the duality because a discussion based on the backward Kolmogorov equation is enough for the aim of the present paper. Hence, we briefly review the key points which are enough to understand the proposals in the next section.

First, we consider the following Fokker-Planck equation instead of the stochastic differential equation \cite{Gardiner2009}:
\begin{eqnarray}\label{eq:original_fokker_planck}
  \frac{\partial}{\partial t} p(\bm{x}, t) = \mathcal{L}p(\bm{x}, t),
\end{eqnarray}
where $p(\bm{x}, t)$ is the probability density function and $\mathcal{L}$ is the time-evolution operator defined as
\begin{eqnarray}\label{eq:time_evol_op_fokker_planck}
  \mathcal{L} = - \sum_i \frac{\partial}{\partial x_i} a_i(\bm{x}) + \frac{1}{2} \sum_{i, j} \frac{\partial^2}{\partial x_i \partial x_j} \left[ B(\bm{x}, t)B(\bm{x})^\mathrm{T} \right]_{i, j}.
\end{eqnarray}
The aim here is to evaluate the $m$-th order moment of the $i$-th element of $x(t)$; i.e., 
\begin{eqnarray} \label{eq:target_statistics}
  M^{(m)}_i(\bm{x}_0,t) = \mathbb{E}_{p(\bm{x}, t|\bm{x}_0, t_0)}[x_i^m],
\end{eqnarray}
where $p(\bm{x}, t|\bm{x}_0, t_0)$ is the probability density function of $\bm{x}(t)$ with the initial condition $\bm{x}(t_0) = \bm{x}_0$. The initial condition is written as
\begin{eqnarray}
p(\bm{x}, t_0) = \delta(\bm{x}-\bm{x}_0),
\end{eqnarray}
where $\delta(\cdot)$ is the Dirac delta function.

Second, we derive the backward Kolmogorov equation instead of the Fokker-Planck equation; as for the backward Kolmogorov equation, see~\cite{Risken_book}. Since the time-evolution of the probability density function is formally denoted as
\begin{eqnarray}
p(\bm{x}, t|\bm{x}_0, t_0) = e^{\mathcal{L}(t-t_0)} p(\bm{x}, t_0),
\end{eqnarray}
the derivation of the time-evolution operator for the backward Kolmogorov equation is easily understood as follows:
\begin{eqnarray} \label{eq:moment_phi}
  M^{(m)}_i(\bm{x}_0,t) & = \int x_i^m p(\bm{x}, t|\bm{x}_0, t_0) d\bm{x} \nonumber                                           \\
                      & = \int x_i^m \left\{ e^{\mathcal{L}(t-t_0)} p(\bm{x}, t_0) \right\} d\bm{x} \nonumber                     \\
                      & = \int x_i^m \left\{ e^{\mathcal{L}(t-t_0)} \delta(\bm{x} - \bm{x}_0) \right\} d\bm{x} \nonumber          \\
                      & = \int \left\{ e^{\mathcal{L}^\dagger (t-t_0)} x_i^m \right\} \delta(\bm{x} - \bm{x}_0) d\bm{x} \nonumber \\
                      & = \int \varphi(\bm{x}, t) \delta(\bm{x} - \bm{x}_0) d\bm{x} \nonumber                               \\
                      & = \varphi(\bm{x}_0, t),
\end{eqnarray}
where $\mathcal L^\dagger$ is the adjoint operator of $\mathcal L$ and $\varphi(\bm x, t)$ is a solution of a time evolution equation
\begin{eqnarray} \label{eq:kolmogorov_backward}
  \frac{\partial}{\partial t} \varphi(\bm{x}, t) = \mathcal{L}^\dagger \varphi(\bm{x}, t).
\end{eqnarray}
The adjoint operator $\mathcal L^\dagger$ is written as
\begin{eqnarray} \label{eq:adjoint_time_evolution_operator}
  \mathcal{L}^\dagger = \sum_i a_i(\bm{x}) \frac{\partial}{\partial x_i} + \frac{1}{2} \sum_{i, j} \left[ B(\bm{x})B(\bm{x})^\mathrm{T} \right]_{i, j} \frac{\partial^2}{\partial x_i \partial x_j},
\end{eqnarray}
which corresponds to the time-evolution operator for the backward Kolmogorov equation. In \eref{eq:moment_phi}, we employed integration by parts and the fact that the probability density functions will vanish at $x_i \to \pm \infty$. Note that the initial condition for \eref{eq:kolmogorov_backward} is
\begin{eqnarray} \label{eq:kol_back_initial_condition}
\varphi(\bm{x}, t_0) = x_i^m,
\end{eqnarray}
which corresponds to the target statistic. Here, we focus on the value of the solution of \eref{eq:kolmogorov_backward} at a certain coordinate $\bm{x}_0$, i.e., $\varphi(\bm{x}_0, t)$; the value immediately gives the expectation of the target statistic, $M^{(m)}_i(\bm{x}_0,t)$. 

Third, we expand $\varphi(\bm x, t)$ with monomial basis functions, $\bm{x}^{\bm{n}} = \prod_i x_i^{n_i}$, so that
\begin{eqnarray} \label{eq:kol_back_power}
  \varphi(\bm{x}, t) = \sum_{\bm{n} \in \mathbb{N}^D} P^{(m)}_i(\bm{n}, t) \bm{x}^{\bm{n}},
\end{eqnarray}
where $\{P^{(m)}_i(\bm{n}, t)\}$ are the expansion coefficients. By substituting this equation into \eref{eq:moment_phi}, we have
\begin{eqnarray} \label{eq:moment_dual}
  M^{(m)}_i(\bm{x}_0,t) = \sum_{\bm{n} \in \mathbb{N}^D} P^{(m)}_i(\bm{n}, t) \bm{x}_0^{\bm{n}}.
\end{eqnarray}
Hence, it is enough to obtain the coefficients $\{P^{(m)}_i(\bm{n}, t)\}$ to evaluate the expectation $M^{(m)}_i(\bm{x}_0,t)$. Employing the basis expansion in \eref{eq:kol_back_power} with the time-evolution equation in \eref{eq:kolmogorov_backward}, we have
\begin{eqnarray} \label{eq:kolmogorov_backward_series_expansion}
  \sum_{\bm{n} \in \mathbb{N}^D} \left\{ \frac{\partial}{\partial t} P^{(m)}_i(\bm{n}, t) \right\} \bm{x}^{\bm{n}} = \mathcal{L}^\dagger \left\{ \sum_{\bm{n} \in \mathbb{N}^D} P^{(m)}_i(\bm{n}, t) \bm{x}^{\bm{n}} \right\},
\end{eqnarray}
which gives the simultaneous ordinary differential equations for $\{P^{(m)}_i(\bm{n}, t)\}$ by comparing the coefficients of the basis expansion. Note that the initial condition should be
\begin{eqnarray}
P^{(m)}_i(\bm{n}, t_0) = \delta_{n_i, m},
\end{eqnarray}
where $\delta_{\cdot,\cdot}$ is the Kronecker delta function. The initial condition stems from $\varphi(\bm{x},t_0)$ in \eref{eq:kol_back_initial_condition}. We will denote examples of the simultaneous ordinary differential equations for $\{P^{(m)}_i(\bm{n}, t)\}$ later, which will help the reader understand the above discussion.

Here are some comments on the above discussion. The first comment is on duality in stochastic processes: the derived equations for $\{P^{(m)}_i(\bm{n}, t)\}$ do not satisfy the law of conservation of probability, so the coefficients $\{P^{(m)}_i(\bm{n}, t)\}$ are not probabilities. As shown in \cite{Ohkubo2013}, it is possible to recover the probabilistic characteristics to extend the discussions, which leads to the connection with the duality relation in stochastic processes. However, as written later, the coefficients $\{P^{(m)}_i(\bm{n}, t)\}$ are enough to learn neural networks; there is no need to recover the probabilistic characters here. The second comment is related to the target statistics. The above discussion is limited to the target statistics in \eref{eq:target_statistics}, i.e., a simple moment. While the restriction simplifies the discussion, we should mention that other statistics, such as correlations, can also be evaluated. 

In summary, the moment $M^{(m)}_i(\bm{x}_0,t)$ is evaluated by solving the simultaneous ordinary differential equations derived from \eref{eq:kolmogorov_backward_series_expansion}, without samplings of stochastic processes. Our goal is to construct a neural network to predict the target moment. The learning procedure needs expectations for various initial coordinates. Furthermore, stochastic cases need many samples for a single initial coordinate to evaluate the expected value, which requires high computational costs. Then, the above approach based on the simultaneous ordinary differential equations enables us to avoid the samplings. However, we still need to prepare a data set, pairs with an input coordinate and the target moment, to use conventional learning frameworks in previous studies. Is there a more efficient method suitable for stochastic differential equations? Next, we propose a simple way to answer this question and directly embed the information in the equations to neural networks via $\{P^{(m)}_i(\bm{n}, t)\}$.

\section{Proposed method}

Let us consider the following neural network with a single hidden layer: The input is the initial coordinate of stochastic process $\bm{x}_0$, the output is an estimate of the moment $M^{(m)}_i(\bm{x}_0,t)$, the number of nodes of the hidden layer is $n$, and the activation function is a sigmoid function $\sigma(x) = 1 / (1+e^{-x})$. While this neural network is not deep, the universal approximation theorem guarantees the approximation ability when we use many hidden nodes \cite{Cybenko1989, Hornik1989}. As demonstrated later, the simple structure works well for examples with nonlinear coefficients. 

The key of the proposed method is the direct comparison of the coefficients $\{P^{(m)}_i(\bm{n}, t)\}$ in \eref{eq:moment_dual} with the corresponding components in the neural network. Although the direct comparison is a simple idea, it yields preferable learning results, as discussed later. 

First, the output of the neural network, $y \in \mathbb{R}$, given the input $\bm{x} \in \mathbb{R}^D$ can be represented as
\begin{eqnarray}
  y &= \bm{q}^\mathrm{T} \bm{\sigma} (R\bm{x} + \bm{s}) \nonumber \\
  &= \sum_{i=1}^n q_i \sigma\left(\sum_{j=1}^D R_{ij}x_j + s_i\right),
\end{eqnarray}
where $\bm{q}, \bm{s} \in \mathbb{R}^n, R \in \mathbb{R}^{n \times D}$ are the weights of the neural network. Using the Taylor expansion of $\sigma(x)$ up to the $N$-th order, we can approximate the output with the power series of the input $\bm{x}$ as follows:
\begin{eqnarray} \label{eq:nn_power_series}
\fl
  y \approx \sum_{\{\bm{l}\in\mathbb{N}^D | \sum_j l_j \le N\}} \sum_{k=\sum_j l_j}^N \sum_{i=1}^n 
\left(
\begin{array}{c}
k \\ 
l_1, \ldots, l_D, k-\sum_j l_j
\end{array}
\right)
\frac{\sigma^{(k)}(0)}{k!} q_i \bm{r}_{i}^{\bm{l}} s_i^{k-\sum_j l_j} \bm{x}^{\bm{l}},
\end{eqnarray}
where $\sigma^{(k)}(\cdot)$ denotes the $k$-th derivative of $\sigma(\cdot)$ and $\bm{r}_i$ is the $i$-th row vector of the weight matrix $R$. Note that a notation for multinomial coefficients 
\begin{eqnarray}
\left(
\begin{array}{c}
k \\ 
k_1, \ldots, k_r
\end{array}
\right)
= \frac{k!}{k_1! \dots k_r!}
\end{eqnarray}
is used in \eref{eq:nn_power_series}.

At this stage, the correspondence between \eref{eq:moment_dual} and \eref{eq:nn_power_series} is clear; both have the basis-expansion form for $\bm{x}^{\bm{n}}$ or $\bm{x}^{\bm{l}}$. Hence, it is possible to employ a direct comparison between them. For example, one can use the sum of the square errors of coefficients in \eref{eq:moment_dual} and \eref{eq:nn_power_series} as a cost function:
\begin{eqnarray} \label{eq:cost_func}
C^{(m)}_{i,t}(\bm{q}, R, \bm{s}) = \sum_{\{\bm{l}\in\mathbb{N}^D | \sum_j l_j \le N\}} \left( P^{(m)}_i(\bm{l}, t) - P_{\mathrm{NN}}(\bm{l}, \bm{q}, R, \bm{s}) \right)^2,
\end{eqnarray}
where
\begin{eqnarray}
\fl
  P_{\mathrm{NN}}(\bm{l}, \bm{q}, R, \bm{s}) = \sum_{k=\sum_j l_j}^N \sum_{i=1}^n 
\left(
\begin{array}{c}
k \\ 
l_1, \ldots, l_D, k-\sum_j l_j 
\end{array}
\right)
\frac{\sigma^{(k)}(0)}{k!} q_i \bm{r}_{i}^{\bm{l}} s_i^{k-\sum_j l_j}.
\end{eqnarray}
By minimizing this cost function $C^{(m)}_{i,t}$, we obtain the neural network that estimates the moment $M^{(m)}_i(\bm{x}_0,t)$. 

Note that the number of simultaneous ordinary differential equations derived from \eref{eq:kolmogorov_backward_series_expansion} is infinite. Of course, we cannot evaluate the infinite number of coefficients $\{P^{(m)}_i(\bm{n}, t)\}$. However, with the comparison in \eref{eq:cost_func}, it is enough to consider a finite set of $\{P^{(m)}_i(\bm{n}, t)\}$ for ${\{\bm{n}\in\mathbb{N}^D | \max_i n_i \le N\}}$; the order of Taylor approximation, $N$, relates to the upper bound. From this fact, it is possible to approximate the simultaneous ordinary differential equations for $\{P^{(m)}_i(\bm{n}, t)\}$ with the limited range of indexes ${\{\bm{n}\in\mathbb{N}^D | \max_i n_i \le N\}}$; we employ this approximation in the next section. Note that it would be natural to determine the order of Taylor approximation from the number of coefficients $\{P_i^{(m)}(\bm{n},t)\}$ used in the cost function in \eref{eq:cost_func}. The coefficients stem from the expansion of the function $\varphi(\bm{x},t)$ in \eref{eq:kol_back_power} for the backward Kolmogorov equation. Hence, the number of coefficients $\{P_i^{(m)}(\bm{n},t)\}$ is related to the approximation accuracy of $\varphi(\bm{x},t)$. In other words, it is enough to use cutoffs that yield enough coefficients $\{P_i^{(m)}(\bm{n},t)\}$ of the function $\varphi(\bm{x},t)$ with the accuracy one wants to approximate.

What kind of features can we expect from the proposed method? Here, we focus on the Taylor-type basis expansion in the proposed method. Hence, one could expect that the approximation performance is better when $\bm{x}_0$ is closer to the origin. Next, we confirm this conjecture with numerical demonstrations.

\section{Numerical examples}

In this section, we demonstrate the proposed method for two stochastic differential equations. The first example is a famous one-dimensional model, i.e., the Ornstein-Uhlenbeck process \cite{Gardiner2009}. The analytical solutions for moments are known for the Ornstein-Uhlenbeck process, and it is easy to compare the performance. The second example is the noisy van der Pol system, which is a two-dimensional model. The noisy version of the original van der Pol system \cite{van_der_Pol1926} has already been used in filtering \cite{Lakshmivarahan2009, Frogerais2012} and recent studies related to data analysis for dynamical systems \cite{Crnjaric-Zic2020}.

In both cases, the simultaneous ordinary differential equations for $P_i^{(m)}(\bm n, t)$ are solved numerically via the functions of the Scipy package, \verb|scipy.integrate.solve_ivp|. The minimization procedure for the cost function \eref{eq:cost_func} is performed with \verb|scipy.optimize.least_squares|, in which we randomly generate the initial values for $\bm{q}, R, \bm{s}$ from a uniform distribution with the range $[-1, 1)$. The initial time $t_0$ is $0$, and the time interval for each case is denoted in table~\ref{tab:params}. Table~\ref{tab:params} also shows the network size, $n$, and the order of Taylor approximation, $N$.

\begin{table}
  \caption{\label{tab:params}%
    Parameter settings used in the two examples.
  }
\begin{center}
  \begin{tabular}{lccc}
\hline\hline
  & Time interval $t$ & \begin{minipage}{35mm} Number of hidden \\ layer nodes $n$\end{minipage} & \begin{minipage}{35mm} Order of Taylor  \\ approximation $N$\end{minipage} \\ 
\hline
      Ornstein-Uhlenbeck  & 1   & 4 & 12 \\
      noisy van der Pol & 0.1 & 8 & 17 \\
\hline\hline
    \end{tabular}
\end{center}
\end{table}

\subsection{Ornstein-Uhlenbeck process}

The stochastic differential equation for the Ornstein-Uhlenbeck process is given as follows \cite{Gardiner2009}:
\begin{eqnarray}
  dx = -\gamma x dt + \sigma dW(t).
\end{eqnarray}
Hence, the time-evolution operator $\mathcal{L}$ for the Fokker-Planck equation in \eref{eq:original_fokker_planck} is given as
\begin{eqnarray}
\mathcal{L} = \frac{\partial}{\partial x} \gamma x + \frac{\partial^2}{\partial x^2} \frac{\sigma^2}{2},
\end{eqnarray}
which leads to the adjoint operator in \eref{eq:adjoint_time_evolution_operator} as
\begin{eqnarray}
\mathcal{L}^\dagger = - \gamma x \frac{\partial}{\partial x} +  \frac{\sigma^2}{2} \frac{\partial^2}{\partial x^2}.
\end{eqnarray}

Since there is only one variable $x$, the coefficients are $\{P^{(m)}_i (n_1, t)\}$ for $n_1 = 0,1,2,\cdots$. The derived simultaneous ordinary differential equations are as follows:
\begin{eqnarray}\label{eq:demo_ou_odes}
\frac{d}{dt} P^{(m)}_i(n_1,t)
= - \gamma n_1 P^{(m)}_i(n_1,t)
+ \frac{\sigma^2}{2} (n_1+2)(n_1+1) P^{(m)}_i (n_1+2, t)
\end{eqnarray}
for $n_1 = 0,1,2,\cdots, N$. Note that we employ the finite cutoff with $N$, and set $P^{(m)}_i (n_1, t) = 0$ for $n_1 > N$. Here, as denoted in Table~\ref{tab:params}, we set $N=12$. 

As for the numerical demonstrations, we construct two neural networks to estimate the first and second-order moments. As for the first-order moment case, the initial condition for \eref{eq:demo_ou_odes} is set as $P^{(m)}_i(n_1,0) = \delta_{n_1,1}$; $P^{(m)}_i(n_1,0) = \delta_{n_1,2}$ is used for the second-order moment case.

The analytical solutions for moments are as follows \cite{Gardiner2009}:
\begin{eqnarray}
  M^{(1)}(x_0,t) & = x_0 e^{-\gamma t},   \\
  M^{(2)}(x_0,t) & = \left(x_0 e^{-\gamma t}\right)^2 + \frac{\sigma^2}{2\gamma}\left(1-e^{-2\gamma t}\right).
\end{eqnarray}

\begin{figure}
  \centering
  \includegraphics[width=140mm]{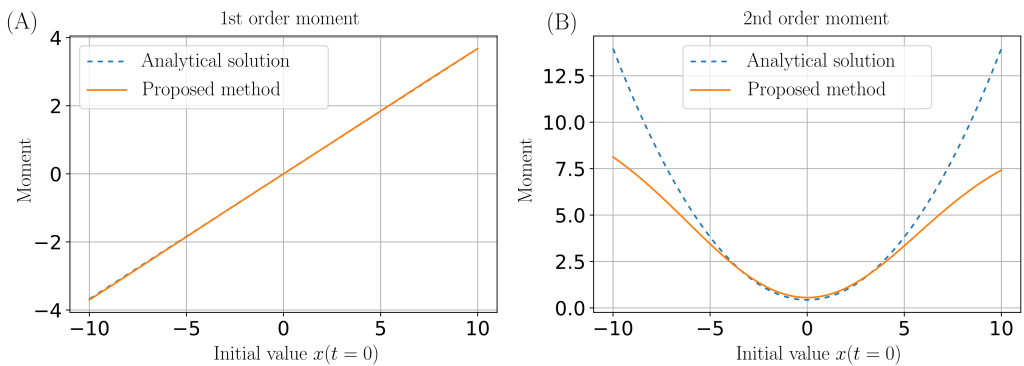}
  \caption{\label{fig:ou_emb_moment}Estimated moments of the Ornstein-Uhlenbeck process by the neural networks constructed with the proposed method. (A) First order moment. (B) Second order moment. The dashed lines correspond to The analytical solutions.}
\end{figure}

Here, we set $\gamma = \sigma = 1$. Figure~\ref{fig:ou_emb_moment} shows the numerical results obtained from the proposed method. We also depict the analytical results. As for the proposed method, after the learning procedures in section~3, various inputs  $x_0 = x(t=0)$ are used to obtain the corresponding outputs, and we drew the curves in figure~\ref{fig:ou_emb_moment}.

The numerical results indicate that the first-order moment is estimated well even in the small neural network. As for the second-order moment, the accuracy decreases as the input value is further away from the origin. This behavior is just the expected one from the characteristics of the Taylor approximations, as discussed in section~3.

We here note that it is possible to change the origin. The expansion of $\varphi(\bm{x},t)$ in \eref{eq:kol_back_power} is around zero. As discussed in \cite{Ohkubo2021}, it is sometimes beneficial to shift the origin, which yields the different values for the coefficients $\{P^{(m)}_i(\bm{n}, t)\}$. Then, the learned networks show good agreements around the shifted origins; we have numerically checked these characteristics. From these results, it is possible to say the idea employed in the backward Kolmogorov equation and the optimization work well.

The Ornstein-Uhlenbeck process is linear. Next, we use a two-dimensional example with nonlinear coefficients.

\subsection{Noisy van der Pol system}

The noisy van del Pol system obeys the following stochastic differential equations:
\begin{eqnarray}
  d
\left(
\begin{array}{c}
x_1\\
x_2 
\end{array}
\right)
= 
\left(
\begin{array}{c}
x_2 \\
 \epsilon x_2(1-x_1^2)-x_1 
\end{array}
\right)
dt + \mathrm{diag}(\nu_{11}, \nu_{22})d\bm{W}(t),
\end{eqnarray}
where $\epsilon$ and $\nu_{11}, \nu_{22}$ are parameters, and $\mathrm{diag}(\cdot)$ means the diagonal matrix with the corresponding diagonal elements.

The corresponding time-evolution operator $\mathcal{L}$ and its adjoint one $\mathcal{L}^\dagger$ are as follows:
\begin{eqnarray}
\mathcal{L} &= - \frac{\partial}{\partial x_1} x_2 - \frac{\partial}{\partial x_2} \left( \epsilon x_2 \left( 1-x_1^2 \right) - x_1\right)
+ \frac{\partial^2}{\partial x_1^2} \frac{\nu_{11}^2}{2}
+ \frac{\partial^2}{\partial x_2^2} \frac{\nu_{22}^2}{2}, \\
\mathcal{L}^\dagger &= x_2  \frac{\partial}{\partial x_1} + \left( \epsilon x_2 \left( 1-x_1^2 \right) - x_1 \right) \frac{\partial}{\partial x_2} 
+ \frac{\nu_{11}^2}{2} \frac{\partial^2}{\partial x_1^2} 
+ \frac{\nu_{22}^2}{2} \frac{\partial^2}{\partial x_2^2}.
\end{eqnarray}
Then, the simultaneous ordinary differential equations for $\{P^{(m)}_i(n_1,n_2,t)\}$ become
\begin{eqnarray}
\fl
\frac{d}{dt} P^{(m)}_i (n_1, n_2,t)
=& 
(n_1+1) P^{(m)}_i (n_1+1, n_2-1,t)
+ \epsilon n_2 P^{(m)}_i (n_1, n_2,t) \nonumber \\
\fl
&- \epsilon n_2 P^{(m)}_i (n_1-2, n_2,t)
- (n_2+1) P^{(m)}_i (n_1-1, n_2+1,t) \nonumber \\
\fl
&+ \frac{\nu_{11}}{2} (n_1+2)(n_1+1) P^{(m)}_i (n_1+2, n_2,t) \nonumber \\
\fl 
&+ \frac{\nu_{22}}{2} (n_2+2)(n_2+1) P^{(m)}_i (n_1, n_2+2,t).
\end{eqnarray}
As in the Ornstein-Uhlenbeck case, a finite cutoff is employed; $P^{(m)}_i (n_1, n_2,t) = 0$ for $n_1 > N$ or $n_2 > N$. 

In the demonstration, we set $\epsilon = 1$ and $\nu_{11} = \nu_{22} = 1$. For the noisy van der Pol systems, there is no analytical result. Hence, we use the estimated values obtained from the method in section~2 as the approximate true value. 

The proposed method is the approach via the right side depicted in figure~\ref{fig:concept}. For comparison, we perform the conventional machine learning approach with backpropagation. As explained in section~1, a naive approach, the left one in figure~\ref{fig:concept}, requires samplings with the Monte Carlo method to estimate the expected values. However, it takes high computational costs. Hence, we here employ the approach via the middle one in figure~\ref{fig:concept}; the expected values for various initial conditions are evaluated with the method in section~2. We make a data set with the pairs of an input coordinate and the target moment. Then, the neural network is learned from the data set. Of course, the Monte Carlo method with a considerably large data set gives the same learned network. The comparison approach requires various initial coordinates for the data set; we generated them from the uniform distribution with the range $[-4, 4]^2$. The data size is $250,000$; the size is chosen to give similar total estimation errors, as explained later. The learning procedure is performed with \verb|PyTorch| with \verb|AdaDelta|. 

Again, note that the proposed method does not need any preparation for the data set. Instead, the coefficients $\{P^{(m)}_i(\bm{n},t)\}$ are numerically evaluated and used.

\begin{figure}
  \centering
  \includegraphics[width=170mm]{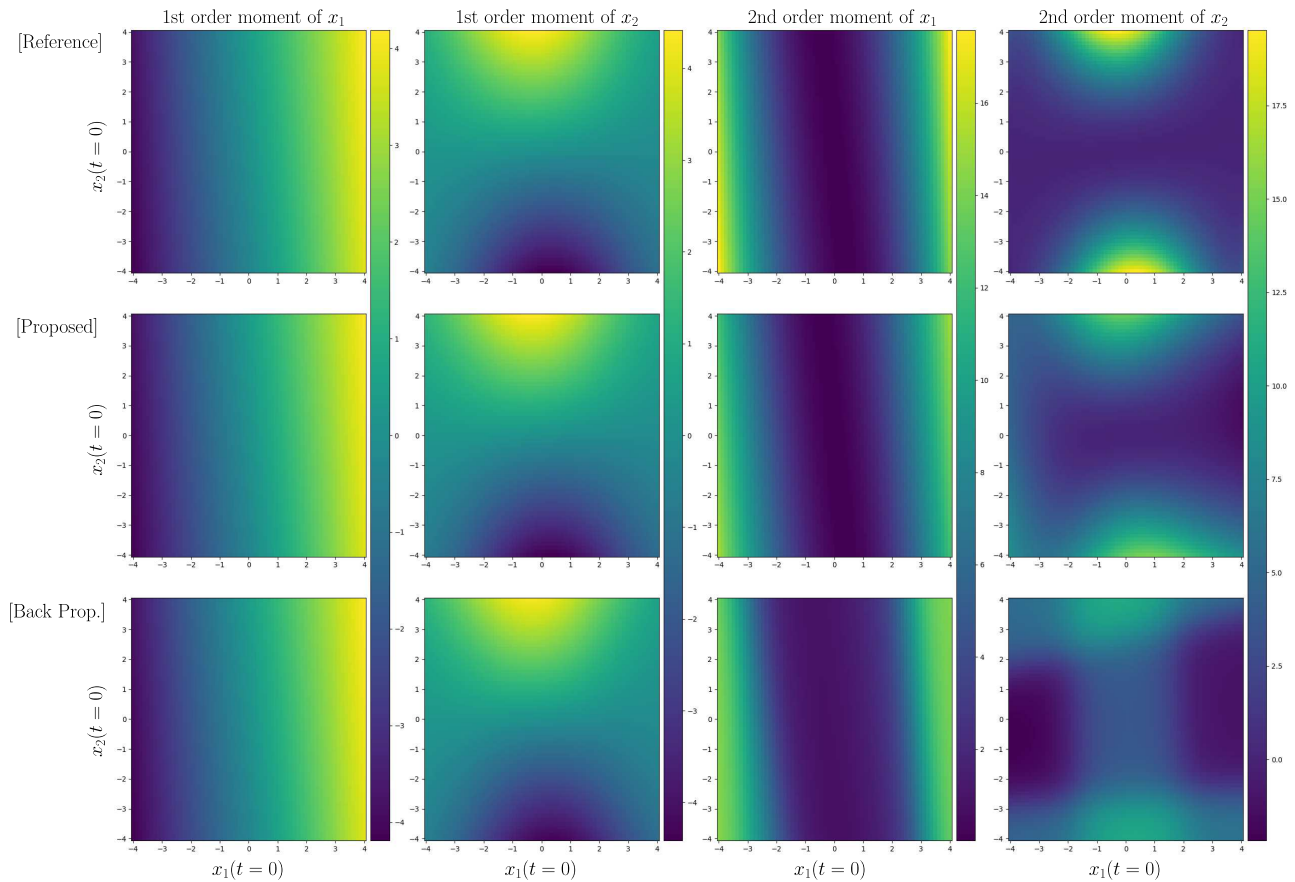}
  \caption{\label{fig:van_learn_moment} Estimated moments for the noisy van der Pol system. (Upper) Approximately true results. (Middle) Results by the proposed method. (Bottom) Results by the conventional learning approach based on backpropagation.}
\end{figure}

The results are shown in figure~\ref{fig:van_learn_moment}. The upper ones correspond to the approximate true results obtained by the method in section~2. The middle ones are the results of the proposed method. The bottom ones correspond to those by the conventional learning approach based on backpropagation. As for the first-order moments of $x_1$ and $x_2$, it is difficult to see the differences among the three cases; the learning results are good enough. As for the second-order moment for $x_2$, we see the differences from the true one. The proposed method gives a similar shape to the true one, while the color is a little thinning away from the origin. On the other hand, the result of backpropagation is worse even near the origin.

\begin{figure}
  \centering
  \includegraphics[width=140mm]{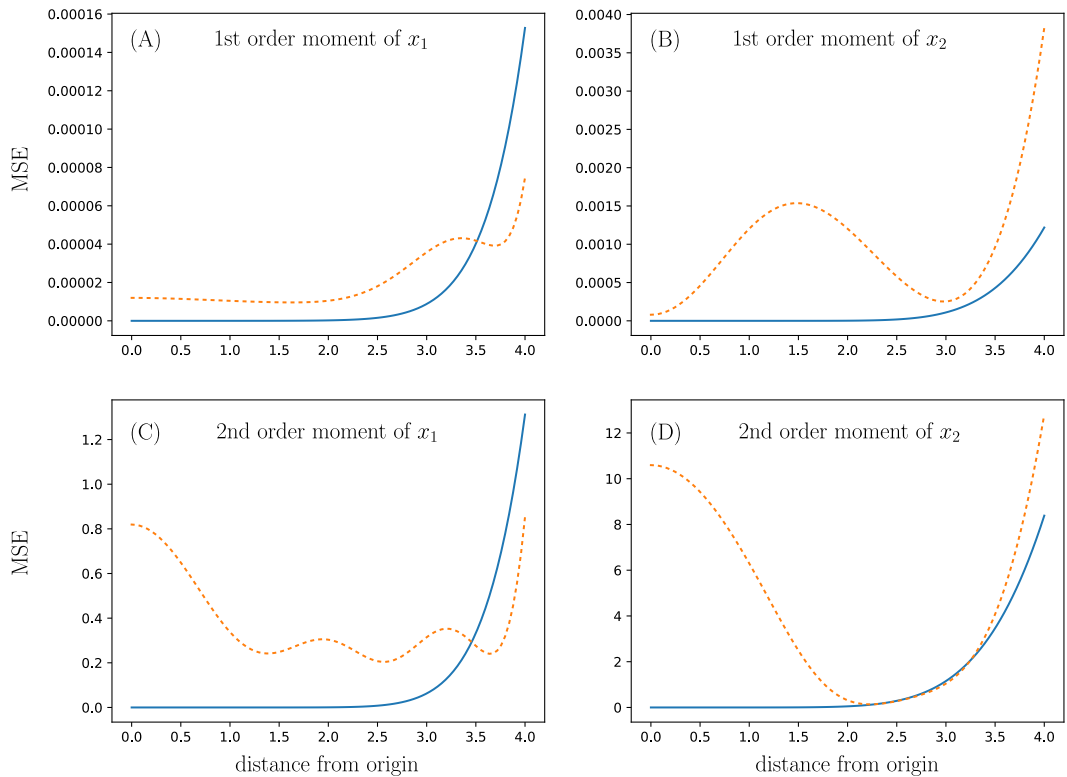}
  \caption{\label{fig:van_acc_comp} The estimated errors. The solid lines correspond to the results for the proposed method, and the dashed lines correspond to those for the backpropagation.}
\end{figure}

To examine this behavior in more detail, we evaluate the mean squared error between the learned results and approximate true ones for the distance from the origin. We calculate the outputs of the learned neural network values for various inputs with the $100 \times 100$ mesh grids for polar coordinates with the range $[0,4] \times [0, 2\pi)$. The results are shown in figure~\ref{fig:van_acc_comp}. Note that the proposed method and the backpropagation method give similar total errors with distances from $0$ to $4$; the data size was chosen so that they would be. Note that we do not intend exactly to yield the same total errors because we here want to focus on the behavior of the errors with respect to the distance. In figure~\ref{fig:van_acc_comp}, we see a clear difference between the results; the proposed method gives more accurate estimations near the origin. This characteristic reflects the feature of the Taylor expansion in the proposed method.

\section{Conclusion}

We proposed a new learning method for neural networks, which target is the statistics of the stochastic differential equations. The proposed method directly compares coefficients obtained from the dual process and the weights in the neural networks. There is no need for sampling procedures for the stochastic processes, and the proposed method gives a different framework from previous ones based on backpropagation. Although the idea is naive, the numerical demonstration shows remarkable features of the proposed approach; we have more accurate estimations near the origin, and the errors increase with distance. This feature stems from the Taylor-type basis expansion in the derivation of the method. Of course, it is easy to shift the origin of the state space using Ito's lemma, as commented in section 4.1. Hence, we can select another coordinate as the origin to estimate the moments around the selected coordinate more accurately. Problems such as overfitting are less likely to occur compared with the conventional learning methods from data sets because the proposed method does not require sampling of stochastic processes. These features will be hopeful in some applications in which one wants to embed the information of system equations into neural networks.

The present work is the first attempt utilizing direct comparisons, and the proposed method and the numerical examples would be enough as the demonstration. Of course, there are remaining works, as follows.

First, the estimations of the proposed method are less accurate except for inputs near the origin. Different cost functions with some focused coordinates could improve it. 

Second, we should seek suitable numerical optimization methods to make the algorithm faster. It is interesting to investigate how the performance changes for neural networks with other structures because we used a simple neural network with one hidden layer in the present work. Although the cases of multiple hidden layers are possible in principle, they require solving rather difficult nonlinear optimization problems. Hence, further development of numerical solvers would be necessary.

Third, practical applications would be investigated. For example, the following situation seems to be the case quite often: One knows the form of the time-evolution equations, but there are some ambiguities about the values of parameters. In such a situation, we can use the proposed method to construct an initial neural network based on the information of the equations; then, we apply additional learning steps to the initialized neural network. These procedures will reduce the size of data sets.

Fourth, we need further work to make the method practical in high-dimensional systems. We confirmed that the algorithm based on combinatorics \cite{Ohkubo2022,Ohkubo2021} can deal with at least four or five-dimensional systems rapidly. Furthermore, it is possible to evaluate the information of moments for systems with several dozen variables using the tensor-train format; see, for example, \cite{Gels2017} for the tensor-train format. Of course, the optimization procedure will take longer computational time, and we need to develop further studies in the research community of optimizations. Although the practical applications for higher-dimensional cases are beyond the scope of the present paper, we hope that the present work will motivate further collaboration between statistical physics and machine learning.

The idea proposed in the present paper is the first step for future work. We hope that the present work will open up a new way to use the information of the time-evolution equation directly to learn neural networks.

\ack
This work was supported by JSPS KAKENHI Grant Number JP21K12045.

\end{document}